\pdfoutput=1

\documentclass[11pt]{article}

\usepackage{EMNLP2023}

\usepackage{times}
\usepackage{latexsym}

\usepackage[T1]{fontenc}

\usepackage[utf8]{inputenc}

\usepackage{microtype}

\usepackage{inconsolata}

\usepackage{enumerate}
\usepackage{enumitem}
\usepackage{mathtools}
\usepackage{amsmath}
\usepackage{amssymb}
\usepackage{microtype}
\usepackage{graphicx}
\usepackage{booktabs} 
\usepackage{enumitem}
\usepackage{caption}
\usepackage{subcaption}
\usepackage{array,multirow}
\usepackage{cleveref}

\newcommand{\bo}[1]{\textbf{#1}}
\newcommand{\bw}[1]{\textit{\textcolor{Periwinkle}{#1}}}

\newcommand{\mypara}[1]{\smallskip\noindent\textbf{#1}\,\,}
\newcommand{\myrepro}[1]{\smallskip\noindent\textit{#1}}

\title{Tiny Transformers Excel at Sentence Compression}

\author{Peter Belcak\\
  ETH Zurich\\
  \texttt{belcak@ethz.ch} \\\And
  Roger Wattenhofer\\
  ETH Zurich\\
  \texttt{wattenhofer@ethz.ch}
}

\begin{document}
\maketitle
\begin{abstract}
It is staggering that words of the English language, which are on average represented by 5--6 bytes of ASCII, require as much as 24 kilobytes when served to large language models.
We show that there is room for more information in every token embedding.
We demonstrate that 1--3-layer transformers are capable of encoding and subsequently decoding standard English sentences into as little as a single 3-kilobyte token.
Our work implies that even small networks can learn to construct valid English sentences and suggests the possibility of optimising large language models by moving from sub-word token embeddings towards larger fragments of text.

 
\end{abstract}

\section{Introduction}
\label{section:introduction}

An average English Wikipedia word can be represented by 5.4 characters (ASCII bytes), with humans experiencing no difficulty in comprehending this representation at speed.
This is in stark contrast with the 24 kilobytes required to represent \textit{sub}-word tokens that may be fed into GPT-3 \cite{brown2020language}, the first language model widely lauded for the plausibility of its outputs.
The scaling factor of 4500 prompts a natural question: could we have language models operating on more efficient representations of their inputs, thus decreasing their training and inference cost?

Recent work on the Funnel-Transformer and hierarchical text transformers \cite{dai2020funnel,nawrot2021hierarchical} proposed to introduce information bottlenecks on the sequence length into the original transformer architecture \cite{vaswani2017attention}.
It found that shortening and subsequently expanding sequences of hidden states throughout the transformer leads to a significant decrease in demands on resources while coming at little cost to the model performance.

We explore a related line of research, investigating whether it is feasible for transformers to operate on inputs that are more information-rich than the apparently oversized token embeddings.
In particular, we ask and answer the following:

\mypara{Question.} \textit{Are transformer models capable of synthesising more condensed representations of text and subsequently decomposing them back with little loss in information?}

The significance of this question is the following: if transformers can perform generalising compression with low loss, there is a hope they could implicitly operate on semantically higher-level units of language (e.g. sentences) with little change to the spirit of the usual pre-training procedures.
This is because to operate on higher-level representations of text one needs to be able to internally decompose the representation into individual parts in order not to miss any semantic payload.
A positive answer would then mean (a) an immediate decrease in the resources necessary to train and employ these models in practice, and (b) could have benefits for the quality of model outputs on the grounds of the transformer being able to focus on sentence cohesion rather than token cohesion.

The next natural and easily discernible unit of language above word is sentence. We aim to answer the above question, and to that end, we show that the following is true:

\mypara{Claim.} Even very small transformers are already capable of compressing sentences into a single token and decompressing them back with little loss.

\mypara{Contributions.} 
We demonstrate that the claim is true by training BERT-like \citep{devlin2018bert} transformers consisting of only 1--3 layers to perform sentence compression on standard English language corpora. We evaluate their accuracy when compressing and subsequently decompressing sentences, showing good reconstruction abilities.
We further give an analysis of the impact of changing model dimension, depth, and decoder input width on the quality of the sentence reconstructions.

\noindent
Our work has numerous implications, namely on the nature of unsupervised language model pre-training, transformer model efficiency, and the possibility of a sentence-level transformer architecture.

\mypara{Unsupervised language learning.}
Masked and causal language modelling as pre-training methods do not distinguish between the internalisation of language rules and the learning of factual knowledge.
Our work empirically shows that transformers can turn information compressed to a single token into a grammatically correct English sentence with as little as two BERT layers.
This indicates that much of the representational power of large language models (frequently in the order of dozens of layers) might be dedicated to the memorisation of factual knowledge and its surrounding contexts.

\mypara{Efficiency of the transformer architecture.}
Our results indicate that as far as capturing semantic information carried by words is concerned, the size of language model token embeddings is unjustified.
It is therefore likely that the need for large token embedding stems from elsewhere in the transformer architecture.

\mypara{Sentence-level transformers.}
Our work suggests that a future language model architecture could consist of a transformer encoding sentences or sentence fragments into single-token embeddings, feeding them into a large, core transformer, and the decompressing the outputs of the core transformer with a decoder before passing the hidden embeddings into a language modelling head.

\medskip \noindent
In the interest of full reproducibility, we make all our code, training setup, and data immediately available (see \Cref{appendix:reproducibility_effort}). A demo is further available through institutional servers\footnote{link anonymised}.

\section{Related work}
\label{section:related_work}

\mypara{Token embedding size.}
In its naive form, every dimension of a token embedding requires 4 bytes.
Prior to the introduction of word2vec, it was ``popular'' to use word embeddings with 50-100 dimensions \cite{mikolov2013efficient}.
Word2vec proposed to increase this to 640, but subsequent work demonstrated that embeddings of dimension 100-300 possess the same desirable properties \cite{pennington2014glove}.
The first transformer used 512 dimensions per token \cite{vaswani2017attention}, which was increased to 768 in GPT \cite{radford2018improving}, BERT-base \cite{devlin2018bert}, and DistilBERT \cite{sanh2019distilbert}; to 1024 in BERT-large; to 1600 in GPT-2 \cite{radford2019language}; to 12288 in GPT-3 \cite{brown2020language} and OPT-175bn \cite{zhang2022opt}; and to 18432 in PaLM \cite{chowdhery2022palm}.
In sum, there is an unbroken trend towards larger token embeddings.
Surprisingly, an early effort to comprehensively optimise pre-trained transformers \cite{sanh2019distilbert} found that ``variations on the token embedding dimension have smaller impact on computation efficiency than variations on other factors''.

\mypara{Neural text compression.}
Existing research has demonstrated success of using neural networks to compress textual data.
\citet{goyal2018deepzip} introduced an RNN-based lossless compressor for sequential data, and demonstrated that it achieves a 20\% compression improvement over Gzip (a standard baseline for text compression) on textual and genomic documents.
Further, \citet{bellard2019lossless} presented a transformer-based compressor performing 50\% better against Gzip on the \texttt{enwik8} benchmark.
In both works, the goal was to compress entire documents rather than sentences or paragraphs, and the minimality of the compression size was the sole optimisation goal. Note also that both works resorted to overfitting to the input data and including the weights with the compressed documents to achieve the desired reduction in compression size.

\mypara{Sentence embeddings.}
A related line of research focuses on producing sentence embeddings that are of high quality for downstream tasks such as semantic textual similarity scoring, semantic search, and paraphrase mining.
This effort is concerned with the utility of the sentence embeddings to downstream tasks rather than reconstructibility into the original sentence.
As a result, majority of its approaches use contrastive methods that are destructive to original information to arrive at useful sentence embeddings \cite{reimers2019sentence,thakur2020augmented}.
\citet{wang2021tsdae} proposed TSDAE, a BERT-sized transformer autoencoder trained with the masked language modelling task, whose encoder can be contrastively finetuned to yield good downstream performance.
None of these works, however, investigate the option of reconstructing original sentences from their embeddings and focus solely on the downstream tasks at hand.
As such, they do not set any baselines for our effort.
Curiously, it has also been found that the downstream performance of representations that are ``dense'' (more compressed) decreases quicker than for ``sparse'' representations when the search space expands \cite{reimers2020curse}.

\mypara{Shortening transformers.}
\citet{dai2020funnel} proposed the Funnel-Transformer architecture, which gradually compresses sequences of hidden states to shorter ones and hence reduces the overall computation cost.
The narrowing-down of the transformer is achieved by pooling the output hidden representations by subsequent layers, while the performance is aided by long-range inter-layer skip connections.
Funnel transformer has been shown to outperform the standard transformer of comparable FLOPs on a wide variety of sequence-level prediction tasks.

More recently, \citet{nawrot2021hierarchical} used an approach similar to \citet{dai2020funnel} to extend their work to autoregressive transformers.
The authors claim that the resulting ``Hourglass'' architecture improves language modeling efficiency, referring to perplexity scores on a Wikipedia dataset.

Our work differs from both of the above.
Instead of shortening the internal sequences of hidden states, we hope to compress elements of text before feeding them into a transformer in a way that the transformer can operate with fewer data of possibly higher semantic value.

\section{Data}
\label{section:data}
We use the sentences of a standard language model pre-training corpus \citep{devlin2018bert,sanh2019distilbert,liu2019roberta} combining the English Wikipedia as of the 1st of March 2022 and BookCorpus \cite{bandy2021addressing}.
The 6.5 million documents are then decomposed into 229.4 million sentences using the Punkt sentence tokeniser of NLTK \cite{bird2009natural}.

Apart from 81\,000 exceptionally long or erroneous entries (amounting to $\sim$0.04\% of the raw dataset), the vast majority of the sentences in our dataset are shorter than 512 characters (about 91 words or 120 tokens).
We therefore considered only entries shorter than 512 characters as we believe them to be better representative of English sentences.
Detailed statistics of the final corpus can be found in \Crefrange{table:raw_dataset_statistics}{table:raw_sentence_statistics} of \Cref{appendix:dataset_statistics}, and \Cref{appendix:dataset_distributions} lists corresponding length histograms.

We split off 1 million sentences for the testing set and use a 96-million subset of the above dataset for training.
We make our data easily accessible\footnote{\textit{link anonymised}}.

\section{Model}
\label{section:model}
We tokenise individual sentences using the uncased variant of the BERT tokeniser \cite{devlin2018bert}, and then feed them into a transformer network providing one sentence per input.

The network under training is a transformer autoencoder formed by stacking two groups of transformer layers without cross-attention on top of each other, appended with a language modelling head.
This is illustrated in \Cref{figure:model_diagram} of \Cref{appendix:model_diagram}.
We refer to the two groups as \textit{encoder} and \textit{decoder}, respectively.
Both encoder and decoder have the same number of layers $\ell \in \left\{1,2,3\right\}$.
The basic configuration of the transformer layers is aligned with that of BERT: token embedding size is $d \in \left\{768,1024,2048\right\}$ (depending on the experiment), the number of attention heads used one of $\left\{12, 16\right\}$ (depending on divisibility), with the remaining parameters (hidden dimension, dropout, activation) set exactly as in BERT \cite{devlin2018bert} for ease of comparison.
The encoder is connected to the decoder only through the embedding of its leading token.
This embedding is listed $m \in \left\{1, 2,4,\infty\right\}$ times in a row, and then padded by the constant $1$ to the length of the input sentence before being fed into the decoder. $\infty$ denotes that the embedding is listed as many times as there are tokens in the input sentence (resulting in no padding being applied).
Note that our architecture is similar to that of TSDAE \citep{wang2021tsdae} but is much smaller (in fact, smallest possible) and differs in the use of noise and the direction of input into the decoder.

The training task is to reconstruct the ground truth sentence under the cross entropy loss on the outputs of the language modelling head.
We train for 1 epoch -- every entry in the dataset is seen only once.
Further parameters and the computational cost are described in \Cref{appendix:reproducibility_effort}.

\begin{table*}[t!]
\scalebox{0.88}{
    \begin{tabular}{r|c|ccc|ccc|ccc|ccc}
    
    \multicolumn{1}{c}{} & \hspace{35pt} $m$ & \multicolumn{3}{c}{$1$} & \multicolumn{3}{c}{$2$} & \multicolumn{3}{c}{$4$} & \multicolumn{3}{c}{$\infty$} \\
    \cmidrule(r){3-5}
    \cmidrule(r){6-8}
    \cmidrule(r){9-11}
    \cmidrule(r){12-14}
    
    \multicolumn{1}{c}{$\ell$} & \hspace{35pt} $d$ & 768 & 1024 & 2048 & 768 & 1024 & 2048 & 768 & 1024 & 2048 & 768 & 1024 & 2048 \\ 
    
    \toprule
        \multirow{2}{*}{1}  & mean      & 76.05 & 77.56 & 83.37 &	79.50 &	83.42 &	84.21 &	73.76 &	83.95 &	86.44 &	81.01 & 82.43 &	\bw{87.49} \\
                            & weighted  & 65.31 & 67.19 & 74.08 &	68.88 &	74.14 &	75.09 &	62.62 &	74.33 &	77.60 &	70.84 &	72.64 &	\bw{79.42} \\
        \cmidrule(r){3-14}
        \multirow{2}{*}{2}  & mean      & 94.45 & 95.81 & 97.28 &	95.33 &	95.55 &	\bw{97.62} &	96.39 &	95.64 &	97.47 &	96.15 &	96.41 &	97.20 \\
                            & weighted  & 87.99 & 90.42 & 93.07 &	89.44 &	89.89 &	\bw{93.53} &	91.10 &	89.92 &	93.41 &	90.88 &	91.43 &	93.08 \\
        \cmidrule(r){3-14}
        \multirow{2}{*}{3}  & mean      & 97.76 & 97.84 & \bo{98.50} &	97.80 &	97.92 &	\bo{98.41} &	97.87 &	97.96 &	\bw{\bo{98.54}} &	97.84 &	97.84 &	\bo{98.42} \\
                            & weighted  & 93.70 & 93.98 & \bo{95.44} &	93.79 &	94.13 &	\bo{95.20} &	93.94 &	94.18 &	\bw{\bo{95.44}} &	94.01 &	93.92 &	\bo{95.36} \\
    \bottomrule 
    
    \end{tabular}
}

\caption{
    The results of the experiments described in \Cref{section:experiments}, in percentage points.
    ``mean'' indicates mean accuracy of token reconstruction taken across all sentences, ``weighted'' the mean accuracy weighted by sentence length.
    \bo{Emphasis} and \bw{emphasis} mark the best performance per $m$ and $\ell$, respectively.
}
\label{table:summary_results}
\end{table*}

\section{Experiments}
\label{section:experiments}
For each experiment run, we evaluate the ability of the given trained model to compress input sentences into a single embedding and then reconstruct the original sequence. For the evaluation metric, we look at per-token accuracy.

\mypara{Preliminary experiments.} Prior to the focused experimentation below, we tried modifying the number of attention heads and the dropout.
We hypothesised that the increased number of attention heads would lead to better results in small transformers due to the transformer being allowed to do a larger number of separate computations in the individual heads without mutual interference. Likewise, we thought the increase in dropout could encourage the network to construct more robust representations of sentences.
Neither turned out to be the case: increasing the number of attention heads and dropout both led to a noticeable deterioration in performance across the board.
Furthermore, we tried introducing masking noise to the inputs and tasking the transformers to perform denoising reconstructions.
We found that this too led to decrease in sentence reconstruction accuracy, which we attribute to the small size of our networks.

\mypara{Experiments on $\ell$.}
As per our goal, we train \textit{small} transformers with both encoders and decoders 1--3 layers in depth, observing the effect of transformer depth on the sentence reconstruction accuracy.
Deeper transformers have more representational power, hence the natural hypothesis that they will be able to internalise more of the frequent sentence constructs.

\mypara{Experiments on $d$.}
For each transformer depth, we experiment with token embeddings ranging from 768 (BERT-base) to 2048 (2x GPT-2).
Larger token embeddings allow for more sentence information to fit into a single token, and they also increase the number of trainable parameters in the feedforward layers of the transformer. Therefore, we expect larger embeddings to lead to better sentence reconstruction scores.

\mypara{Experiments on $m$.}
Our preliminary experimentation showed that the number of times the embedding of the first token coming from the encoder is listed before being put into the decoder has a noticeable impact on the quality of the sentence reconstructions.
We therefore also investigate the effect of $m$ on the sentence reconstructions.

\section{Discussion}
\label{section:discussion}
The summary quantitative results are listed in \Cref{table:summary_results}, with the relationship to sentence length and a qualitative analysis given in \Crefrange{appendix:reconstruction_accuracy}{appendix:qualitative_analysis}.

\mypara{More layers help.} We observe that consistently across all parameters, the accuracy of the reconstructions improves with the increasing transformer depth $\ell$.
Note also that the jump in accuracy is much more significant on $\ell=1\to2$ than on $\ell=2\to3$.

\mypara{Bigger embeddings lead to marginal improvements.} We find that the effect of increasing embedding size on the quality of reconstructions is consistently positive but mostly marginal (1-3 ppts.) and of significance only for $\ell=1$.

\mypara{Decoder input multiplier plays a role.}
We see that increasing $m$ improves the accuracy, but not always.
The effect of increasing $m$ is most easily spotted for $\ell=1$, but note that for most configurations of $\ell,d$ the optimal $m$ is $2$ or $4$.

\mypara{Compression performance for $\ell=2,3$ is high.}
We see that the best 2--3-layer configurations can compress and subsequently decompress an average sentence with 98\% accuracy, and that over the whole test datasets 95\% of individual tokens are correctly reconstructed at their ground-truth position within the sentence.

\smallskip
\noindent
Considering the diversity of the data (various Wikipedia articles and books), the size of the encoder networks considered (8-25\% of BERT/GPT bodies for $d = 768, \ell \in \left\{1,2,3\right\}$, cf. \Cref{appendix:model_sizes}), and the fact that the sentences are compressed into a single token embedding, this is an extraordinarily good performance in the context of the growing token embedding size in language modelling.
We mention the implications and make the comparisons in \Crefrange{section:introduction}{section:related_work}.

\newpage
\section{Limitations}
\label{section:limitations}

Our work focuses on demonstrating that it is viable for transformers to compress entire sentences into vectors corresponding to a single input token, and that this can be achieved even with very small transformers.
Our experimentation supports this claim.

While we do study how the loss of compression depends on several parameters of the architecture, we do not investigate whether the resulting representations can be feasibly used for downstream tasks.
This is an intentional limitation given the scope of this format.
We believe it possible that the compressed sentence representations obtained by the minimalistic approach will actually perform signficantly worse than the larger sentence embeddings produced by large transformer architectures tailored specifically to that end.

We note that the datasets used in our study consist mostly of well-formed English sentences and give results specific to English.
It is possible that the compression of sentences of less-structured language (such as transcripts of spontaneous speech) would be more difficult to achieve.

We do not include hidden feedforward layer width as a parameter of our experimentation -- since there is a well-documented interplay between width of hidden layers and depth of the network \cite{goodfellow2016deep}, we keep the hidden layer width fixed and vary only the depth of the network.
This ensures and also limits the direct interpretation of the first implication of \Cref{section:introduction} to BERT, DistilBERT, and GPT models.

\bibliography{anthology,custom}
\bibliographystyle{acl_natbib}


\clearpage
\appendix

\section{Dataset details}
\label{appendix:dataset_details}

\subsection{Statistics}
\label{appendix:dataset_statistics}
\Cref{table:raw_dataset_statistics} gives the document-level statistics of the datasets used for the training and testing of our models.
\Cref{table:raw_sentence_statistics} gives the sentence-level statistics per dataset used.

\subsection{Distributions}
\label{appendix:dataset_distributions}
\begin{figure}[h]
    \centering
    \includegraphics[width=1\linewidth]{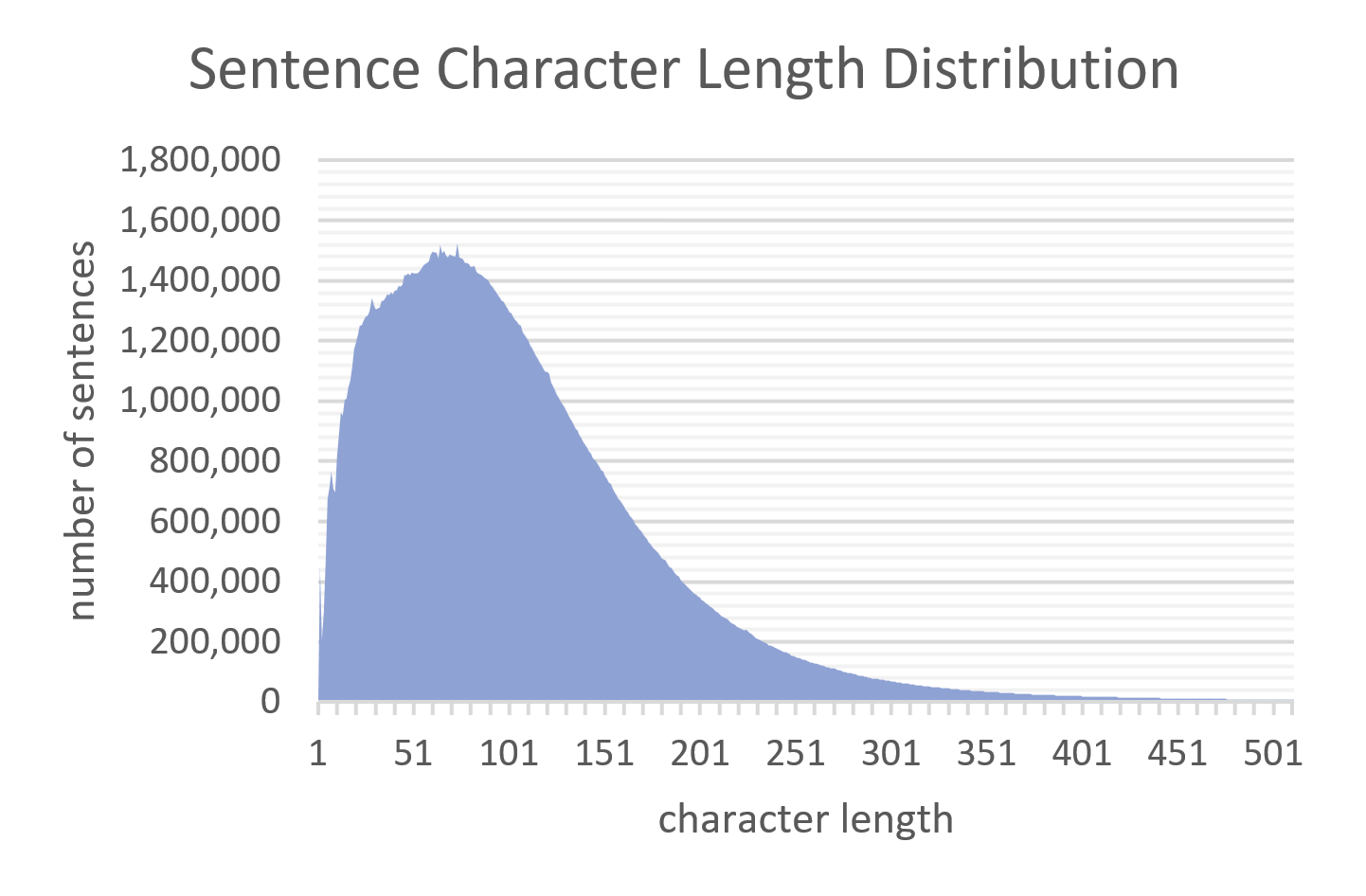}
    \caption{The distribution of sentence lengths in characters in the \textit{combined} corpus. The horizontal axis shows the sentence length in characters, the vertical axis shows the number of sentences in the resulting corpus having that length.}
    \label{figure:chars}
\end{figure}

\begin{figure}[h]
    \centering
    \includegraphics[width=1\linewidth]{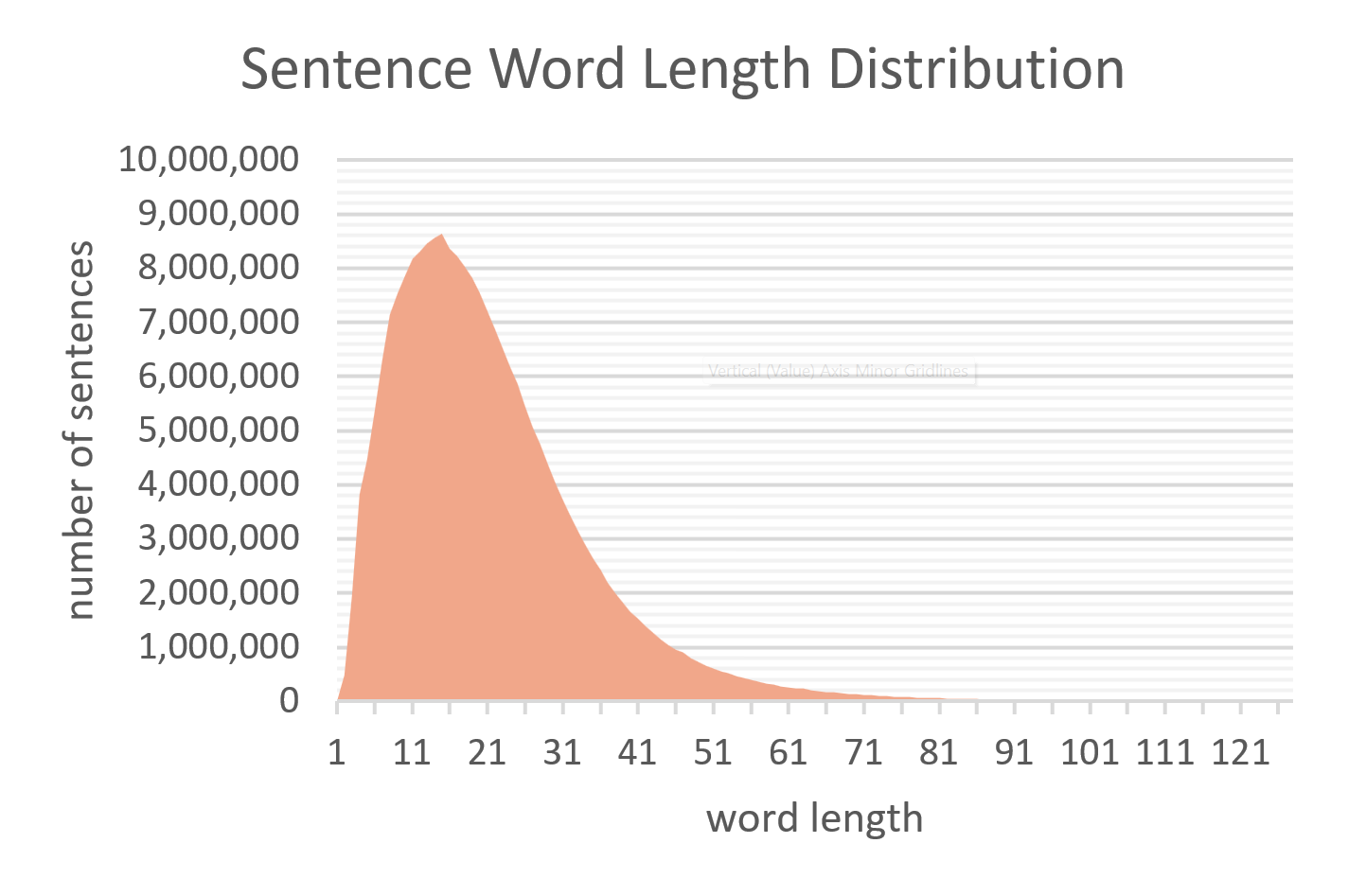}
    \caption{The distribution of sentence lengths in NLTK words in the \textit{combined} corpus. The horizontal axis shows the sentence length in words after tokenising with NLTK, the vertical axis shows the number of sentences in the resulting corpus having that length.}
    \label{figure:words}
\end{figure}

\begin{figure}[h]
    \centering
    \includegraphics[width=1\linewidth]{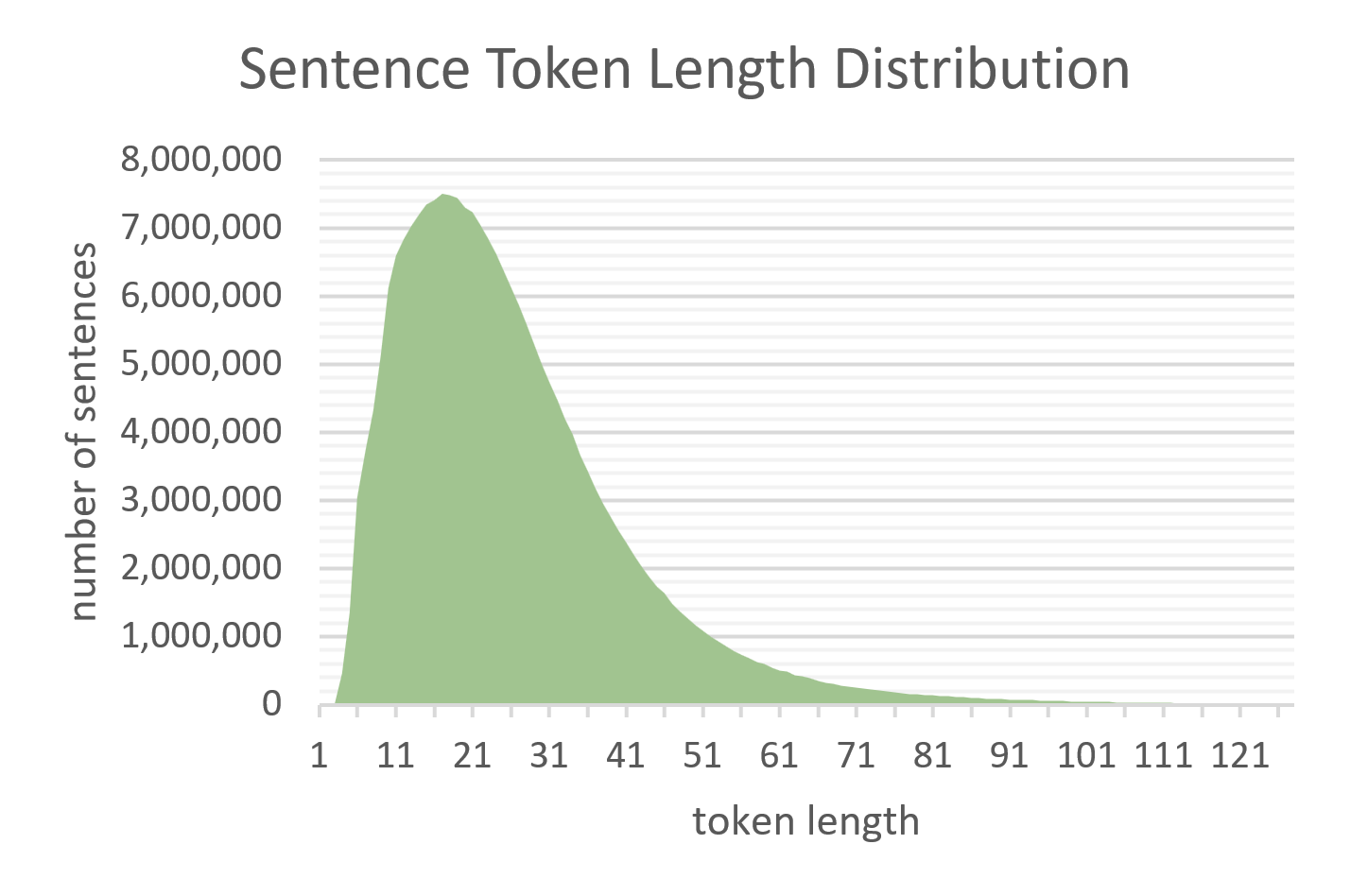}
    \caption{The distribution of sentence lengths in uncased BERT tokens in the \textit{combined} corpus. The horizontal axis shows the sentence length in words after tokenising with the BERT tokeniser, the vertical axis shows the number of sentences in the resulting corpus having that length.}
    \label{figure:tokens}
\end{figure}

\Crefrange{figure:chars}{figure:tokens} show the distribution of sentences in the resulting (combined) corpus as a function of character, word, and token lengths.
We observe that each distribution appears to follow a Gamma distribution, and that very long sentences (>70 words) are a clear rarity in the data.

\begin{table*}[h!]
\centering
\scalebox{1.0}{
    \begin{tabular}{l|rrr}
    
    \toprule
     Dataset \,\,  & \multicolumn{1}{c}{\textsc{Wiki}} & \multicolumn{1}{c}{\textsc{BCO}} & \multicolumn{1}{c}{\textit{combined}} \\
    \midrule
        \multirow{1}{*}{\# of documents}        &\, 6\,458\,670       &	17\,868         &   6\,476\,538   \\

        \multirow{1}{*}{\# of sentences}\,\,\,  &\, 136\,547\,563     &	90\,921\,888    &   227\,469\,451  \\
    \midrule
        \multirow{1}{*}{\# of characters}       &\, 17\,337\,280\,117  &	6\,373\,876\,249 &	23\,711\,156\,366  \\

        \multirow{1}{*}{\# of words}            &\, 3\,225\,292\,505   &	1\,400\,517\,565 &	4\,625\,810\,070  \\

        \multirow{1}{*}{\# tokens}              &\, 4\,023\,124\,041   &	1\,718\,789\,836 &	5\,741\,913\,877  \\
    \midrule
        \multirow{1}{*}{characters per word}    &\, 5.38            &	4.55          &	5.13  \\

        \multirow{1}{*}{characters per token}   &\, 4.31            &	3.71          &	4.13  \\
        
        \multirow{1}{*}{tokens per word}        &\, 1.25            &	1.23          &	1.24  \\
    \bottomrule 
    
    \end{tabular}
}

\caption{
    A summary of document-level statistics of the datasets used (after filtering out 81\,000 overly long or erroneous entries).
    ``Token'' and ``word'' denote BERT-uncased token and NLTK word as in \Cref{section:data}. 
}
\label{table:raw_dataset_statistics}
\end{table*}

\begin{table*}[h!]
\centering
\scalebox{1.0}{
    \begin{tabular}{l|rrr|rrr|rrr}
    
    \toprule
     Dataset \,\,  & \multicolumn{3}{c}{\textsc{Wikipedia}} & \multicolumn{3}{c}{\textsc{BookCorpusOpen}} & \multicolumn{3}{c}{\textit{combined}} \\
\midrule
    
    \multicolumn{1}{c}{} &
    \parbox[t]{3mm}{\multirow{3}{*}{\hspace{-22pt}\rotatebox[origin=c]{45}{characters}}} &
    \parbox[t]{3mm}{\multirow{3}{*}{\hspace{-15pt}\rotatebox[origin=c]{45}{words}}} &
    \parbox[t]{3mm}{\multirow{3}{*}{\hspace{-18pt}\rotatebox[origin=c]{45}{tokens}}} &
    \parbox[t]{3mm}{\multirow{3}{*}{\hspace{-22pt}\rotatebox[origin=c]{45}{characters}}} &
    \parbox[t]{3mm}{\multirow{3}{*}{\hspace{-15pt}\rotatebox[origin=c]{45}{words}}} &
    \parbox[t]{3mm}{\multirow{3}{*}{\hspace{-18pt}\rotatebox[origin=c]{45}{tokens}}} &
    \parbox[t]{3mm}{\multirow{3}{*}{\hspace{-22pt}\rotatebox[origin=c]{45}{characters}}} &
    \parbox[t]{3mm}{\multirow{3}{*}{\hspace{-15pt}\rotatebox[origin=c]{45}{words}}} &
    \parbox[t]{3mm}{\multirow{3}{*}{\hspace{-18pt}\rotatebox[origin=c]{45}{tokens}}} \\
    \multicolumn{10}{c}{} \\
    \multicolumn{10}{c}{} \\
    
    \toprule
        \multirow{1}{*}{mean}           &\, 127.0   &	23.6  &	29.5  \,&\,	70.1 &	15.4 &	18.9  \,&\,	104.2   &	20.3 &	25.2 \\

        \multirow{1}{*}{stddev}\,\,\,   &\, 76.2    &	13.6  &	16.5  \,&\,	54.8 &	10.6 &	12.5  \,&\,	73.9    &	13.1 &	15.9 \\

    \midrule
        \multirow{1}{*}{median}         &\, 113     &	31    &	26    \,&\,	57 &	13   &	16    \,&\,	90    &	18    & 22 \\

        \multirow{1}{*}{$Q(25\%)$}      &\, 74      &	14    &	18    \,&\,	31 &	8    &	11    \,&\,	51    &	11    &	14 \\

        \multirow{1}{*}{$Q(75\%)$}      &\, 163     &	30    &	37    \,&\,	94 &	20   &	24    \,&\,	140   &	27    &	32 \\

        \multirow{1}{*}{IQR}            &\, 89      &	16    &	19    \,&\,	63 &	12   &	13    \,&\,	89    &	16    &	18 \\
    \midrule
        \multirow{1}{*}{$Q(95\%)$}      &\, 271     &	49    &	60    \,&\,	173 &	35  &	41    \,&\,	243   &	45    &	54 \\

        \multirow{1}{*}{$Q(99\%)$}      &\, 309     &	56    &	68    \,&\,	200 &	40  &	47    \,&\,	279   &	51    &	62 \\
    \bottomrule 
    
    \end{tabular}
}

\caption{
    A summary of sentence-level statistics for the datasets used.
    ``Token'' and ``word'' denote BERT-uncased token and NLTK word as in \Cref{section:data}.
    $Q(N\%)$ denotes the Nth percentile, ``IQR'' the inter-quartile range, and ``stddev'' the standard deviation.
}
\label{table:raw_sentence_statistics}
\end{table*}

\section{Model diagram}
\label{appendix:model_diagram}
\begin{figure}[h!]
    \centering
    \includegraphics[width=0.90\linewidth]{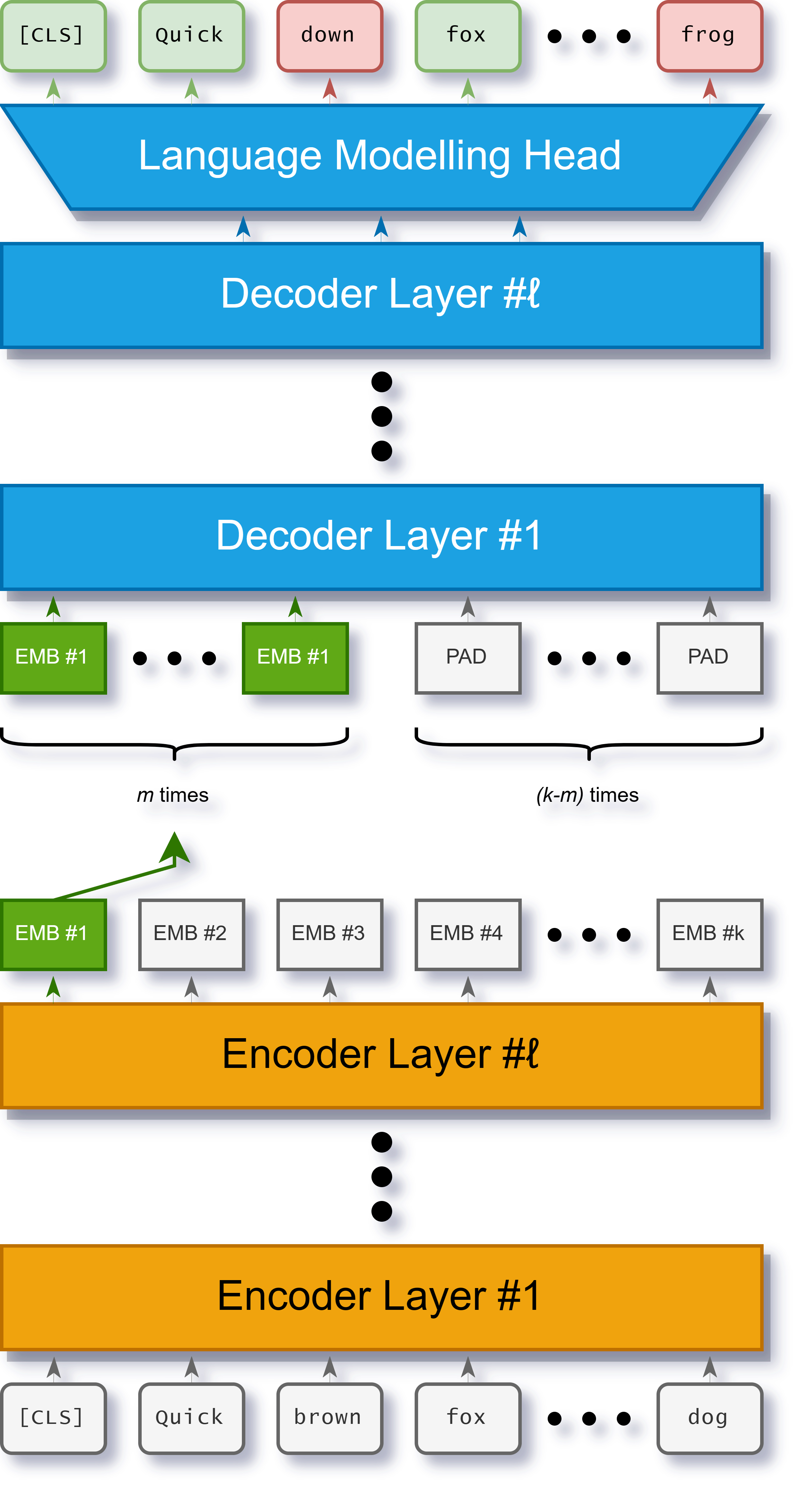}
    \caption{A diagram of the model described in \Cref{section:model}.}
    \label{figure:model_diagram}
\end{figure}

\Cref{figure:model_diagram} shows the diagram of the model used to perform the sentence compression.

\section{Reconstruction Accuracy as a Function of Sequence Length}
\label{appendix:reconstruction_accuracy}

\begin{figure}[t!]
    \centering
    \includegraphics[width=1\linewidth]{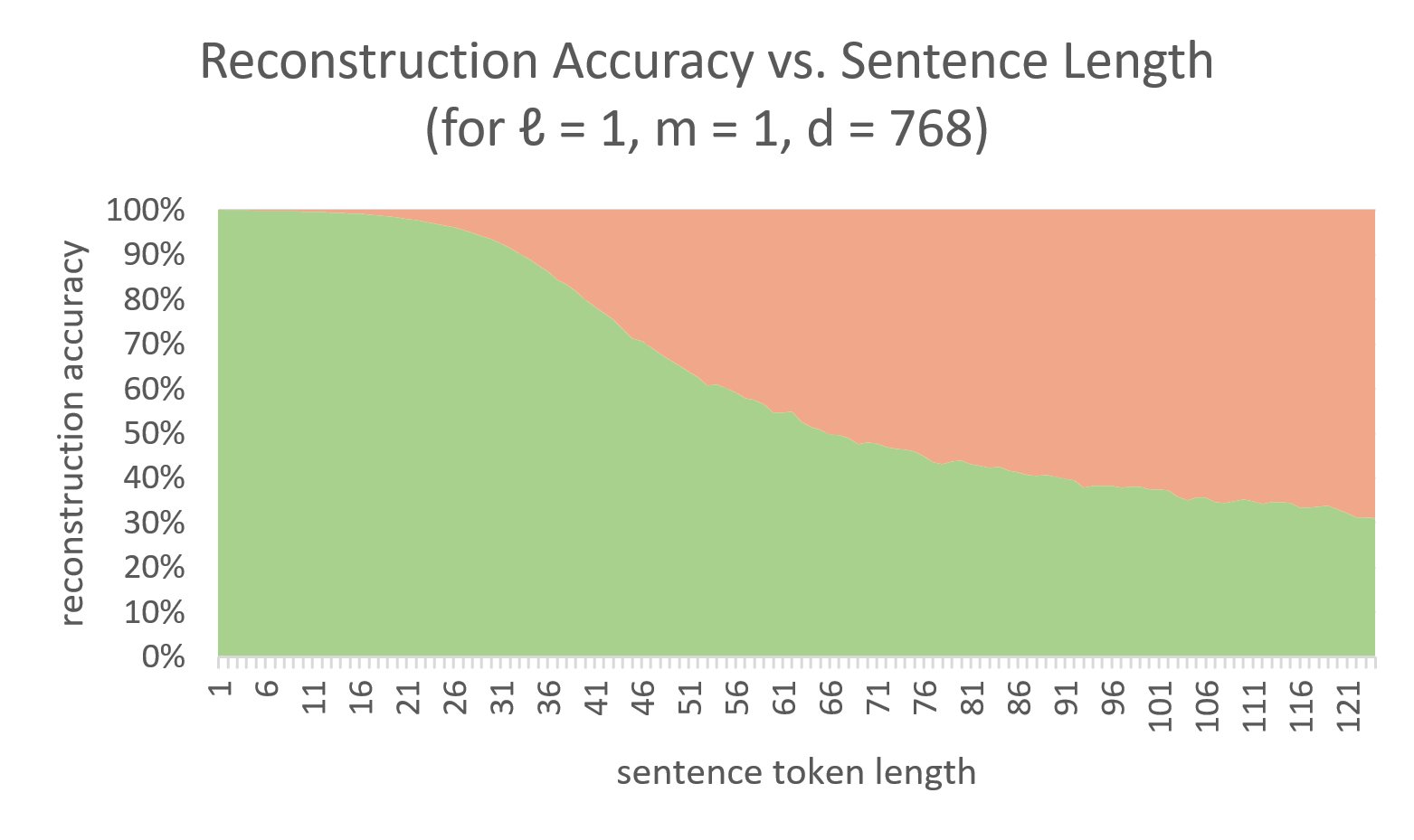}
    \caption{The reconstruction accuracy of a model with $\ell=1,m=1,d=768$ on the test set plotted against the token length of test sentences. The horizontal axis shows the sentence length in tokens, the vertical axis shows the mean reconstruction accuracy for that length.}
    \label{figure:rasl1}
\end{figure}

\begin{figure}[t!]
    \centering
    \includegraphics[width=1\linewidth]{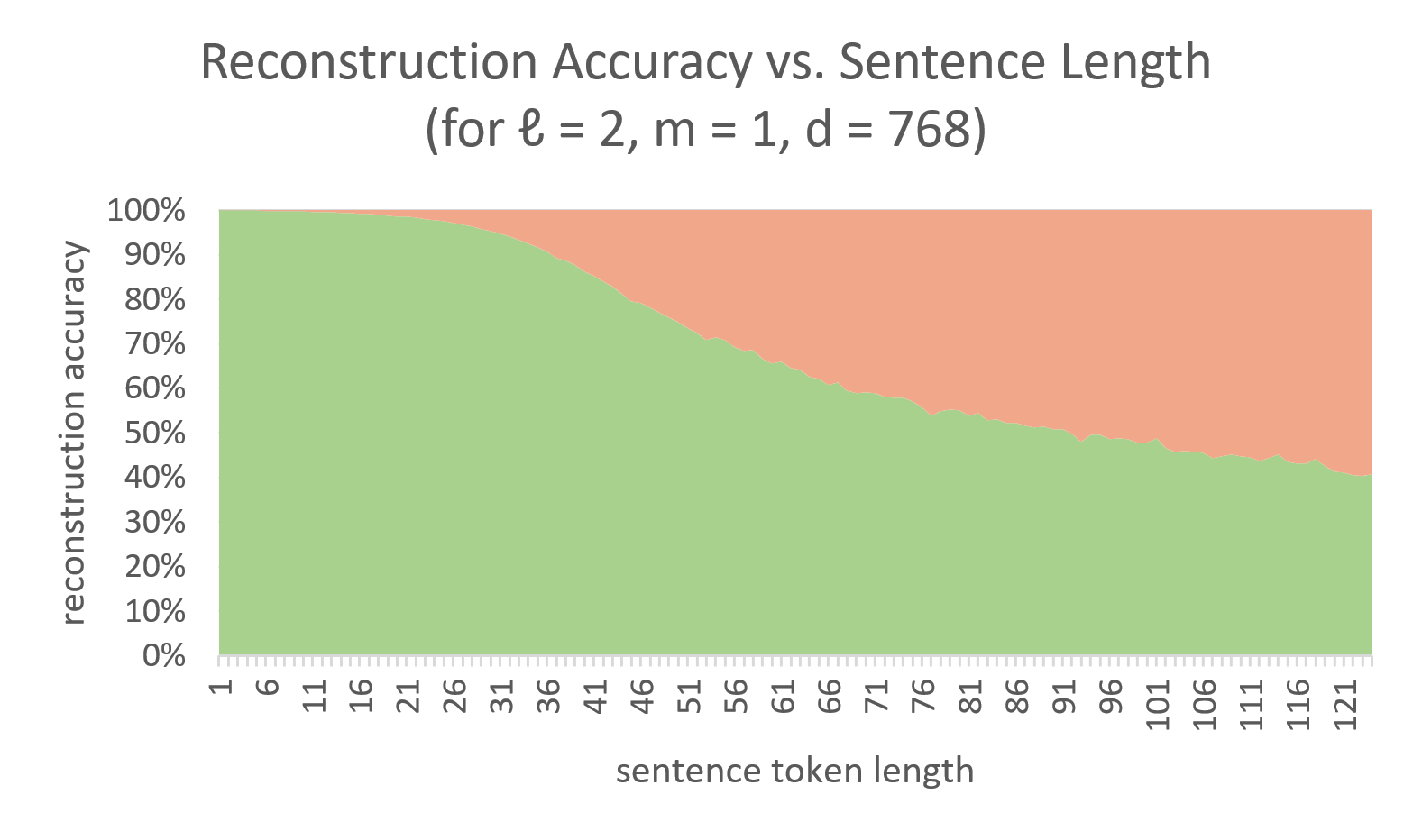}
    \caption{The reconstruction accuracy of a model with $\ell=1,m=1,d=768$ on the test set plotted against the token length of test sentences. The horizontal axis shows the sentence length in tokens, the vertical axis shows the mean reconstruction accuracy for that length.}
    \label{figure:rasl2}
\end{figure}

\begin{figure}[t!]
    \centering
    \includegraphics[width=1\linewidth]{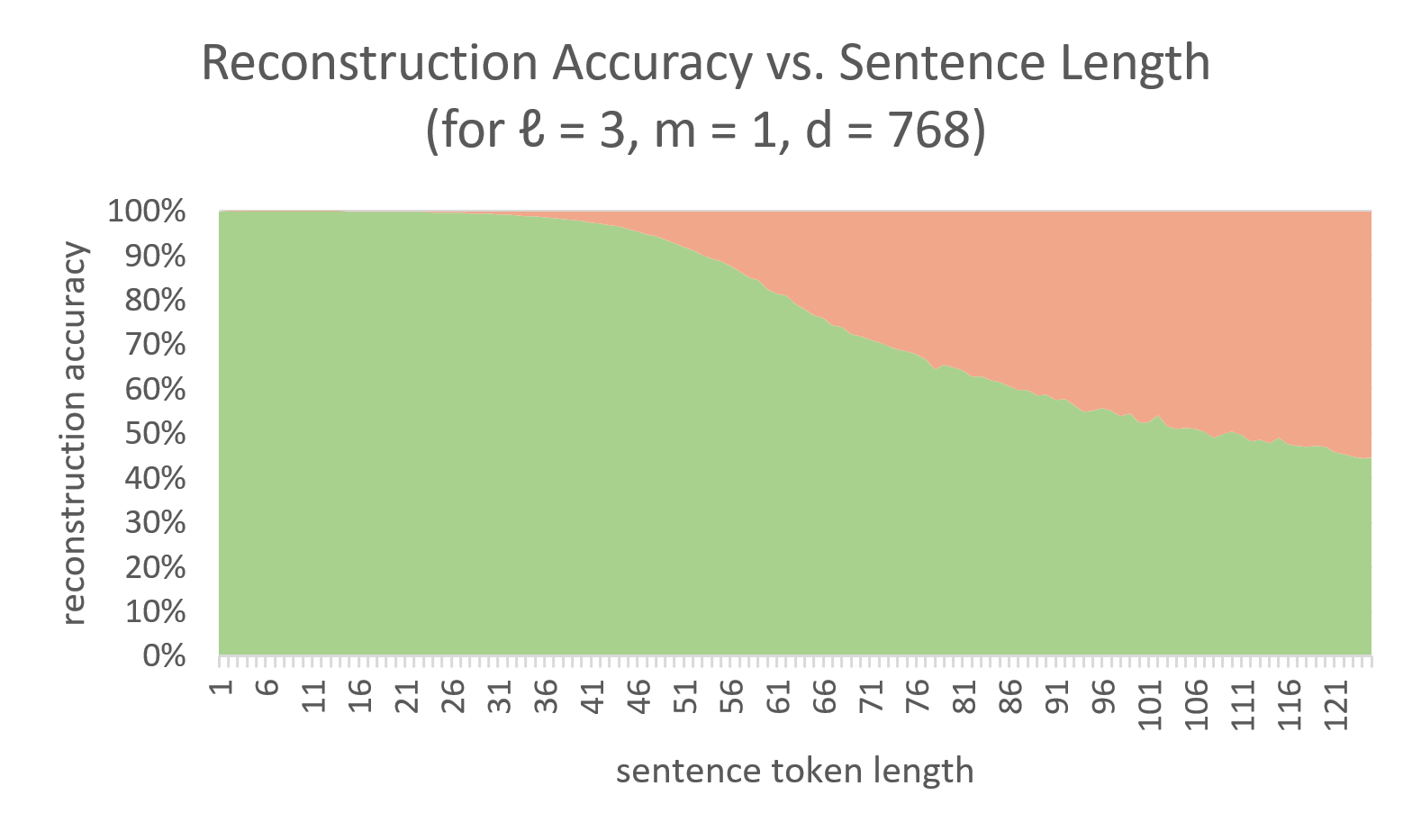}
    \caption{The reconstruction accuracy of a model with $\ell=1,m=1,d=768$ on the test set plotted against the token length of test sentences. The horizontal axis shows the sentence length in tokens, the vertical axis shows the mean reconstruction accuracy for that length.}
    \label{figure:rasl3}
\end{figure}

To investigate the nature of sentence reconstruction errors, we plotted the reconstruction accuracy against the sentence length in tokens.
\Crefrange{figure:rasl1}{figure:rasl3} show the results for a range of models with embedding multiplier $m=1$ and token embedding size $d=768$.
We observe that the increase in layer depth extends the range of sentence lengths that are reconstructed with very high (>95\%) accuracy, before the accuracy begins to fall.

We note, however, that increases in transformer depth also lead to the increase in the reconstruction accuracy for the longest of sequences, and that even for the one-layer compression transformer, the accuracy stands at about $35-40\%$.

\begin{figure}[h]
    \centering
    \includegraphics[width=1\linewidth]{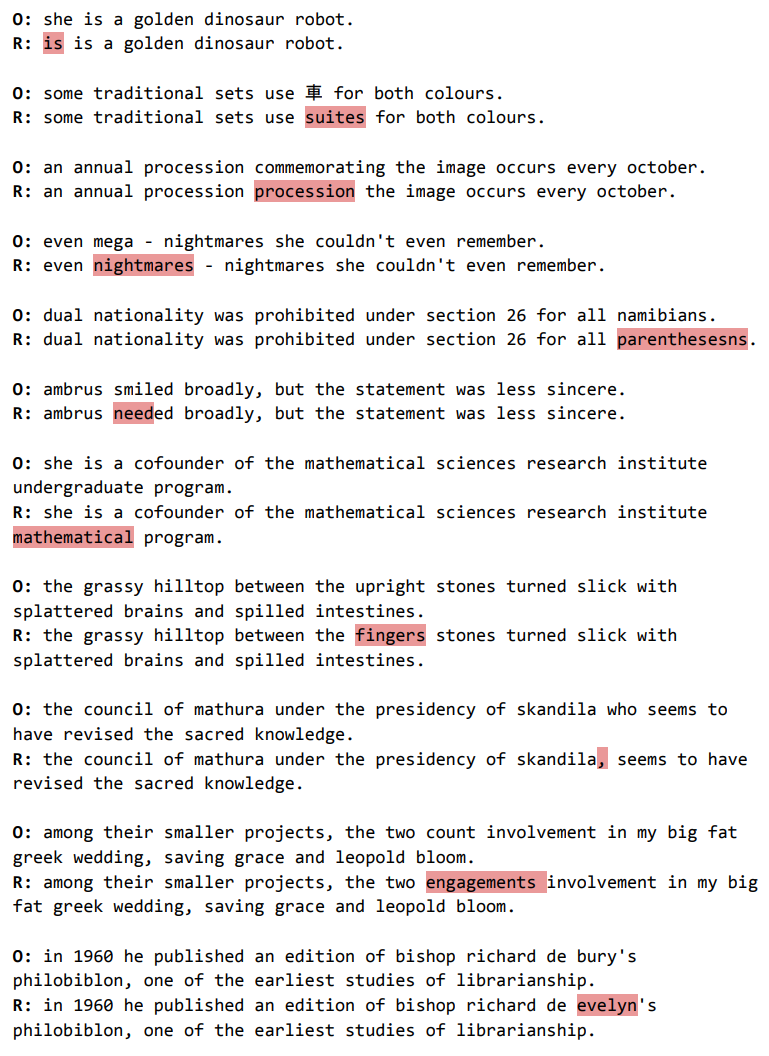}
    \caption{Examples of original sentences and their reconstructions, randomly selected from examples of token length between 10 and 30 and containing at least one erroneous token prediction. The model used had $\ell=3,m=1,d=768$. \textbf{O} denotes the original sentence, \textbf{R} the result of the reconstruction after passing through the model. Background indicates incorrectly predicted token(s).}
    \label{figure:examples}
\end{figure}

\section{Qualitative Analysis of Sentence Reconstructions}
\label{appendix:qualitative_analysis}

\Cref{figure:examples} gives examples consisting of original sentences and their reconstructions by a model that achieves an overall $\sim$97\% reconstruction accuracy.
We observe that many incorrectly predicted tokens are simply a repetition of a token that has previously occured in the sentence.
Furthermore, we find that while models do not struggle with reconstructing terms or expressions that are clearly rare in the language corpus, they might make an occasional mistake if too many such expressions appear in a single sentence, suggesting that the limits on the information content of the embeddings are being hit.
In such cases, the tokens predicted tend to be tokens that more statistically plausible in the language as a whole.

\section{Reproducibility effort}
\label{appendix:reproducibility_effort}
The following has been compiled according to the EMNLP 2023 reproducibility criteria.

\subsection{Experimental results}

\myrepro{A clear description of the mathematical setting, algorithm, and/or model} is provided in \Cref{section:model}, \Cref{appendix:model_diagram}, and references to related models are provided where relevant.

\myrepro{Submission of a \texttt{.zip} file containing source code, with specification of all dependencies, including external libraries, or a link to such resources (while still anonymized)} is provided through OpenReview.
All seeds and other parameters are meticulously noted and easily configurable through a console-line interface.
The link to the anonymised repository is \texttt{https://anonymous.4open.science/r/CAE2}.

\myrepro{A live demo} is further available through institutional servers (\textit{link anonymised}).

\myrepro{Description of computing infrastructure used.}
We run all our experiments on NVIDIA RTX 3090 GPUs (allocating one GPU per experiment).
Our code is written using the PyTorch library.

\myrepro{Training parameters.} 
We train with batch size of 128 (simulated through batch size of 16 and 8 gradient accumulation steps).
Our training is performed with Adam optimiser, with learning rate $1e-4$ if $d=768$ and $5e-5$ for $d=1024,2048$.
We leave all other parameters to the optimiser's PyTorch defaults.

\myrepro{Model parameter counts} are given in \Cref{appendix:model_sizes}.

\myrepro{The average runtime for each model or algorithm (e.g., training, inference, etc.).}
We find that our training takes 36-48 hours to complete (depending on the model size), and that testing on the test dataset requires between 25 and 40 minutes.

\myrepro{Explanation of evaluation metrics used.} Both our loss (cross entropy) and main quality metric (reconstruction accuracy) are classical to the field.

\myrepro{We do not perform hyperparameter searches as a part of our reported experimentation.}

\subsection{Datasets}

\myrepro{Relevant details such as languages, and number of examples and label distributions, as well as details of train/validation/test splits} are given in \Cref{section:data} and the references to the original dataset publications.

\myrepro{Explanation of any data that were excluded, and all pre-processing steps} are also given in \Cref{section:data}.

\myrepro{Our data} consists of the English Wikipedia and BookCorpusOpen datasets, both freely available from multiple sources.
Furthermore, we make our particular processed data and splits readily available through a popular platform at \textit{anonymised}.

\section{Model sizes}
\label{appendix:model_sizes}

\Cref{table:model_sizes} lists the sizes of the various parts of the models used.
Note that the vocabulary size is fixed by the uncased BERT tokeniser, but that the sizes of token embedding tables and language modelling heads vary depending on the size of the token embedding, and that both constitute a major part of the final parameter count.

BERT-base (110 million parameters, $\ell=12, d=768$), RoBERTa-base (125 million, $\ell=12, d=768$),  BERT-large (345 million, $\ell=24, d=768$), and GPT (117 million, $\ell=12, d=768$) models can be directly compared with the sizes in the first column.
There, the bodies of our transformers are between $\frac{1}{12}$ and $\frac{1}{4}$ for 12-layer transformers and correspondingly half that for the 24-layer variants.

\begin{table}[h!]
\centering
\scalebox{0.90}{
    \begin{tabular}{l|rrr}
    
    \toprule
     Model part \,\,\hspace{50pt} $d$  & \multicolumn{1}{c}{$768$} & \multicolumn{1}{c}{$1024$} & \multicolumn{1}{c}{$2048$} \\
    \midrule
        \multirow{1}{*}{token embedding table}      &\, 23.44   &	31.25     &   62.51   \\
    \midrule
        \multirow{1}{*}{body with $\ell = 1$}\,\,\, &\, 5.36    &	7.41      &   17.04  \\
    
        \multirow{1}{*}{body with $\ell = 2$}       &\, 10.72   &	14.82     &   34.09  \\

        \multirow{1}{*}{body with $\ell = 3$}       &\, 16.08   &	22.23     &	  51.13  \\
    \midrule
        \multirow{1}{*}{language modelling head}    &\, 23.44   &	31.25     &	  62.51 \\
    \bottomrule 
    
    \end{tabular}
}

\caption{
    A summary of parameter counts of different parts of the models used, in millions.
    ``Body'' denotes the body of encoder/decoder transformers positioned between the token embedding table on the input size and language modelling head on the output side of the model.
}
\label{table:model_sizes}
\end{table}

\end{document}